\documentclass{article}

\usepackage{arxiv}

\usepackage[utf8]{inputenc} % allow utf-8 input
\usepackage[T1]{fontenc}    % use 8-bit T1 fonts
\usepackage{hyperref}       % hyperlinks
\usepackage{url}            % simple URL typesetting
\usepackage{booktabs}       % professional-quality tables
\usepackage{amsfonts}       % blackboard math symbols
\usepackage{nicefrac}       % compact symbols for 1/2, etc.
\usepackage{microtype}      % microtypography

\usepackage{subcaption}
\usepackage{multirow}
\usepackage{graphicx}
\usepackage[numbers]{natbib}

\title{Active learning for medical code assignment}

\author{
  Martha Dais Ferreira\\
  Dalhousie University\\
  Faculty of Computer Science\\
  \texttt{daismf@dal.ca} \\
   \And
 Michal Malyska \\
  Semantic Health\\
  University of Toronto\\
  \texttt{michal@semantichealth.ai} \\
  \And
 Nicola Sahar \\
  Semantic Health\\
  \texttt{nick@semantichealth.ai} \\
  \And
 Riccardo Miotto \\
  Semantic Health\\
  Icahn School of Medicine at Mount Sinai\\
  \texttt{riccardo.miotto@mssm.edu} \\
  \And
 Fernando Paulovich \\
  Dalhousie University\\
  Faculty of Computer Science\\
  \texttt{paulovich@cs.dal.ca} \\
  \And
 Evangelos Milios \\
  Dalhousie University\\
  Faculty of Computer Science\\
  \texttt{eem@cs.dal.ca} \\
}

\begin{document}
\maketitle

\begin{abstract}
Machine Learning (ML) is widely used to automatically extract meaningful information from Electronic Health Records (EHR) to support operational, clinical, and financial decision making. However, ML models require a large number of annotated examples to provide satisfactory results, which is not possible in most healthcare scenarios due to the high cost of clinician labeled data. Active Learning (AL) is a process of selecting the most informative instances to be labeled by an expert to further train a supervised algorithm. We demonstrate the effectiveness of AL in multi-label text classification in the clinical domain. In this context, we apply a set of well-known AL methods to help automatically assign ICD-9 codes on the MIMIC-III dataset. Our results show that the selection of informative instances provides satisfactory classification with a significantly reduced training set (8.3\% of the total instances). We conclude that AL methods can significantly reduce the manual annotation cost while preserving model performance.
\end{abstract}

% keywords can be removed

%--------------------------------------------------------------
\section{Introduction}
%--------------------------------------------------------------

Extraction of clinical information from unstructured data to support operational, clinical, and financial decision making is typically manually conducted based on heuristics and previous knowledge and expertise. This manual process leads to an inefficient way of delivering care to patients in the health system~\cite{miotto2020identifying}. In this context, many researchers have been studying and applying various Machine Learning (ML) techniques to automatically extracting valuable insights from this unstructured data and extract diagnosis and procedure insights from clinical notes and images~\cite{figueroa2012active, Hohman2019, lee2020biobert, Stevens2019RepresentingDS}. ML techniques performed well on various feature extraction, classification, and regression tasks~\cite{Goodfellow2016, Budd2019ASO, figueroa2012active}.

One issue of the application of ML models is that they require a considerable amount of labeled and stratified training data to ensure robust learning, which is not always possible to obtain in the medical domain that usually lacks labeled datasets, contains rare events, and is difficult to access due to privacy~\cite{holzinger2016interactive, wang2018comparison, von2011statistical}. The lack of labeled data occurs because the annotation costs are high since domain experts are needed to manually label the data (like doctors, nurses, or health administrators).

The interaction of an expert human in the learning process is known as human-in-the-loop machine learning (HILML), in which the expert interacts with the training process to enhance the information extraction and the optimization performance. This approach can reduce the exponential search in the functions space and feature selection~\cite{holzinger2016interactive, Hohman2019}. HILML techniques allow the expert to analyze and explore the data space and the features extracted based on the impact of changes in the prediction, identifying findings and insights that can support and improve the ML performance~\cite{Hohman2019, xin2018accelerating}.

In healthcare, a common HILML used is Active Learning (AL), which has been applied to clinical text data to reduce the annotation cost, i.e., selecting more relevant notes that require minor effort to be annotated~\cite{chen2015study, kholghi2016active, li2019efficient, feder2020active, ji2019cost}. AL is typically used when the training dataset is small due to the difficulty to label samples, requiring a human annotator along the learning process to improve the model~\cite{Budd2019ASO, chen2012applying, figueroa2012active, holzinger2016interactive, kholghi2016active}. Currently, Deep Learning models have been providing good results in the medical domain on clinical text. However, most studies involving AL with Deep Learning models have focused on image datasets, leaving the case of text as an open problem~\cite{chen2012applying, figueroa2012active, gal2017deep, kholghi2016active, wang2016cost}.

Thus, this research aims to evaluate the possibility of employing AL methods in the medical domain, focusing on coding assignment the clinical notes. Therefore, we focus on the application of the most common active learning methods in the clinical notes of MIMIC-III database under the discharge summaries category. In this scenario, the goal is automatically to assign the ICD-9 (the Ninth Revision of International Classification of Diseases) codes based on the content of the notes. It is worth mentioning that, due to the domain aspects, the data is imbalanced and follows a power-law distribution with respect to the number of samples that correspond to each class. Besides, the assignment of ICD-9 codes is a multi-label task, i.e., the sample can be associated with one or more labels or classes~\cite{mullenbach2018explainable, huang2019empirical}. Such aspects motivate this research, because AL methods can select the most informative samples to be labeled, reducing the annotation cost, and provide a better data balance for the training set.

Besides, AL can support data exploration and model interpretation, bringing information for the user about the model design and data that present uncertainty. In that sense, researchers in the visual analytics field are proposing methods that can support the interpretation of ML models in order to provide a better understanding of the data and model behavior, supporting an exploratory analysis~\cite{chen2012applying, Gehrmann2020VisualIW, Hohman2019, vellido2019importance}. These tools can be combined with Active Learning models to facilitate the interaction of non-ML users with the system, improving the interpretability of AI/HIML systems in medicine.

%--------------------------------------------------------------
\section{Related Work}
%--------------------------------------------------------------

Some approaches have been proposed to support automatically assign the ICD-9 codes using clinical notes of MIMIC-II and MIMIC-III, in which deep learning proves to be the most efficient technique~\cite{chen2015study, mullenbach2018explainable, huang2019empirical, ji2020dilated}. Initially, machine learning methods were applied to tackle the assignment of ICD-9 codes, in which the use of hierarchical and flat Support Vector Machines (SVM) for MIMIC-II database was proposed by \citet{perotte2014diagnosis}, who concluded that the hierarchical nature of ICD-9 codes improves the multi-label classification. In sequence, deep recurrent networks outperform other machine learning methods applied in this context as provided by \cite{nigam2016applying}, whose results indicate that recurrent networks capture better information as vector representations. Also, the authors claimed that a better study of the medical vocabulary could improve the assignment of ICD-9 codes.

Therefore, Convolutional Neural Networks (CNNs) started to be applied in literature, improving the results in the medical domain~\cite{li2018automated, mullenbach2018explainable, ji2020dilated}. In this context, the DeepLabeler, proposed by \citet{li2018automated}, is composed of a CNN with doc2vec, showing that the convolution operation was the most effective component. Therefore, researchers started to include different operations in the CNN, such as attention mechanism and dilated convolutions, in order to improve the predictions~\cite{mullenbach2018explainable, ji2020dilated, ji2020dilated}.

To provide a benchmark for the prediction of ICD-9 codes in the MIMIC-III database, deep learning, and machine learning methods were evaluated by \citet{huang2019empirical}, who analyzed different combinations between two vectorization methods and six different classifiers. In summary, all these approaches require a large amount of labeled data to ensure effective learning of the machine and deep learning models~\cite{von2011statistical}. However, code assignment of clinical notes is performed manually, which is expensive, time-consuming, and inefficient. Thus, two potential strategies for dealing with a small and/or imbalanced training set: i) semi-supervised learning, which considers the data distribution for propagating label information through neighboring samples; and ii) Active Learning (AL), which selects the most informative samples for a human agent to label next based on given criteria~\cite{fu2013survey}.

In the medical domain, it is essential to include the domain expert (e.g., medical coder) in the process in order to provide a more accurate coding of the clinical notes. This approach motivates the use of Human-in-Loop strategies, which can be combined with AL methods, once the specialist can provide the labels of the most informative samples selected~\cite{Budd2019ASO, chen2015study, holzinger2016interactive}. The selection of the most informative samples can reduce the manual cost to label the samples and also provide a good representation of the data distribution, supporting the classifier model to achieve a useful generalization, leading to less complex structures~\cite{demello2020shattering, demello2017providing}.

In this context, a survey about AL for named entity recognition in clinical text was provided by \citet{chen2015study}, in which they confirm that AL methods reduce the annotation cost for each sentence. Such reduction was also confirmed by other researchers, whose evaluated AL methods in different tasks, such as classification of clinical text, identification of possible personally identifiable or sensitive sentences, and identification of breast cancer~\cite{kholghi2016active, li2019efficient, feder2020active}. In summary, the models combined with AL methods reduced around $60\%$ of labeled documents required to provide satisfactory results, and $35\%$ of annotation time performed by a specialist~\cite{ji2019cost, kholghi2016active}.

The sample selection is typically conducted based on an acquisition function that measures the uncertainty of the model when predicting unlabeled samples~\cite{chen2015study, fu2013survey, li2019efficient, ji2019cost}. If the model produces a high uncertainty for a particular instance, it means that the model has not enough knowledge to classify that instance, concluding that such an instance is good to incorporate in the training set~\cite{fu2013survey}. Thus, the design of the acquisition function to evaluate the most informative samples to be labeled is the main challenge in the AL field. However, depending on the domain, an issue with the uncertainty metrics is that noise and outliers can be selected, jeopardizing the learning process of the model. Therefore, another strategy to select the most informative samples is to apply feature correlation derived from the semi-supervised learning algorithms. Thus, the idea is that grouping the samples based on their similarities can support the selection of a representative instance for a subset of samples, avoiding outliers instances~\cite{fu2013survey}.

In the case of deep learning algorithm, some other approaches were developed. For instance, the Learning Loss AL includes a loss prediction module in the learning model to predict the loss of unlabeled samples \citet{yoo2019learning}. Another DL technique is the Bayesian Active Learning by Disagreement (BALD) developed by \citet{houlsby2011bayesian}. Some studies concluded that BALD does not perform well with an imbalanced test set and in the first few samples~\cite{houlsby2011bayesian, kirsch2019batchbald}.

Finally, in the medical domain, a deep learning algorithm combined with AL strategies was applied to tackle the problem of medical symptoms recognition from patient text by \citet{mottaghi2020medical}, whose employed clustering methods, selecting the samples closest to the centroid. They conclude that such a strategy can balance the training set, selecting equally samples associated with different labels. In this context, to the best of our knowledge, AL strategies applied in the medical domain for coding assignment have not been deeply explored, motivating this study to evaluate the performance of common AL strategies combined with well-known classifiers applied in MIMIC-III dataset.

%--------------------------------------------------------------
\section{Methodology}
%--------------------------------------------------------------

In supervised machine learning, algorithms must induce some function $f : \mathcal{X} \rightarrow \mathcal{Y}$ that minimizes the classification/regression error for a given learning task, producing a classifier in some space $\mathcal{F}$ (algorithm bias). Such algorithms consider a training set composed of independent samples that are selected from the joint probability distribution $P(\mathcal{X} \times \mathcal{Y})$. Each instance is defined as $(x_i,y_i) \in (\mathcal{X},\mathcal{Y})$, in which $x_i$ is a vector that represents the attributes and $y_i$ is its correspondent label for an $i$ instance. Typically, an instance is associated with a single label, which is the case of binary and multi-class classification. However, in the context of the coding assignment, each instance is associated with one or more labels, known as multi-label classification. In this scenario, $y$ is also represented by a vector containing the label's information.

The databases for such a task commonly have a small number of labeled clinical documents due to the cost of manually labeling them by the medical specialists. Also, these databases are imbalanced. Therefore, the database contains more samples for some labels than others. In fact, in  the MIMIC-III dataset used in this evaluation, the data is highly unbalanced, containing no examples for almost $50$\% of ICD-9 codes. Such an imbalanced and small number of labeled clinical notes can jeopardize the learning in supervised algorithms. These issues motivate the application of Active Learning (AL), which typically receives a set of labeled instances $D_l = {(x_0,y_0),(x_1,y_1),\cdots, (x_n,y_n)}$ and huge number of unlabeled instances $D_u = {(x_0,y_0),(x_1,y_1),\cdots, (x_m,y_m)}$.

Thus, the goal of AL is to select the most informative instances to be manually label by a specialist and be incorporated in the training set, because, in most cases, the training set has not enough data to learn each label. This selection can avoid the time consuming of manual annotation and provide a good representation of the joint probability distribution $P(\mathcal{X} \times \mathcal{Y})$, i.e., a stratified and enough samples to conduct the learning. Acquisition functions (e.g., evaluation metrics) are applied to measure the relevance of instances in $D_u$ to select the most informative one. The typically two strategies applied are uncertainty metric and correlation, in which the former evaluates how much the model is uncertain about a particular instance, while the latter measures the similarity among instances~\cite{Budd2019ASO, fu2013survey}. Therefore, we applied Least Confidence (LC) and Binary Entropy measures as uncertainty metrics and kmeans++ as a feature correlation strategy in the MIMIC-III database to evaluate the performance of AL in the coding assignment.

%--------------------------------------------------------------
\subsection{Frameworks}
%--------------------------------------------------------------

When working with text data, the first step as data mining is to convert the text into numeric vectors. As we are following the \citet{huang2019empirical}'s frameworks, we used Term Frequency–Inverse Document Frequency (TFIDF) and Word2Vec with Continuous Bag-of-Words Model (CBOW) in this study. TFIDF attempts to represent the importance of words in the document~\cite{aggarwal2012introduction}. Such a method extracts words that appear often in a document. However, it is not common in the collection under analysis. TFIDF is commonly used as a feature extractor in the MIMIC database, which is followed by a machine learning technique in order to perform the coding assignment~\cite{nigam2016applying, huang2019empirical, perotte2014diagnosis}. On the other hand, Word2Vec employs neural networks to convert words into numeric vectors keeping the semantic relationship among them~\cite{mikolov2013efficient}.

In the coding assignment of the MIMIC database, Support Vector Machine (SVM), Logistic Regression, and Random Forest are the most employed machine learning techniques. However, they were outperformed by deep learning models, such as Feed-forward Neural Network (FNN), Convolutional Neural Networks (CNN), and Recurrence Neural Networks (RNN). Typically, the multi-label task seems to be a binary classification problem, in which a model is created for each ICD-9 code under analysis. Then, the model will classify if the clinical note is associated or not with this particular code. Thus, in machine learning models, a model is trained for each label, while, for deep learning models, each neuron in the output layer will represent a label and provide a probability for each one.

A benchmark to serve as a baseline for new approaches for this task was proposed by \citet{huang2019empirical}, which we will use to evaluate the efficiency of AL in the coding assignment. They employed Logistic Regression and Random Forest, in which Logistic Regression provided the best results. Logistic regression estimates the probability of samples belonging to a class or not, a binary classification task~\cite{bishop2006pattern}. On the other hand, Random Forest is an ensemble learning method that combines the classification provided by a set of non-correlated decision trees in order to avoid overfitting and reduce the variance~\cite{friedman2001elements}.

In the context of deep learning, they employed Convolutional Neural Network (CNN), Feed-Forward Neural Network (FNN), Long Short-Term Memory Neural Network (LSTM), and Gated Recurrent Unit Neural Network (GRU). So, we started employing the FNN, which is also known as Multilayer Perceptron (MLP), using backpropagation to update the parameter of the network~\cite{huang2019empirical, nigam2016applying}. One challenge faced by researchers is to identify the best design of the network to produce a good classification~\cite{bishop2006pattern}. In this aspect, \citet{huang2019empirical} employed a FNN with $3$ hidden layers, having $5000$, $500$, and $100$ neurons respectively, while \citet{nigam2016applying} employed $2$ hidden layers with $300$ and $100$ neurons respectively. In both cases, they employed ReLU as activation function in hidden layers, sigmoid function in the output layer, binary cross-entropy as the loss function, and stochastic gradient descent as the optimizer. In case of top $100$ ICD-9 codes, \citet{nigam2016applying} uses $1000$ and $1000$ neurons. 

%--------------------------------------------------------------
\subsection{Active Learning methods}
%--------------------------------------------------------------

In this scenario, we started employing the uncertainty metric, in which the model predictions are used to evaluate the unlabeled instances, whose predicted results are represented by vectors that contain the posterior probability of each label. We applied Binary Entropy and Least Confidence measures as acquisition functions to select the most informative sample. Least confidence measures the uncertainty of instances using the higher posterior probability produced by the model among all labels under analysis. Thus, in the case of multi-class classification, the highest posterior probability provided the most likely label that the instance should be associated with. However, we are dealing with multi-label classification that is transformed into a binary classification. Then, if the posterior probability provided is closest to $0.5$, the model is uncertain about the classification of that instance~\cite{fu2013survey}. It is worth to mention that, when the task is multi-label classification, a threshold of the probability is set according to the domain under analysis~\cite{Budd2019ASO, fu2013survey}. The Least Confidence $LC(x_k)$ measure of an instance $k$ is defined in equation~\ref{eq:LC}, in which $p_{\theta}(y|x_k)$ corresponds to the posterior probability of instance $x_k$ to belongs to class $y$. The closer the Least Confidence measure is to $0$, the more uncertain the model is about that particular instance.

\begin{equation}\label{eq:LC}
LC(x_k) = |0.5 - p_{\theta}(y|x_k)|
\end{equation}

The entropy introduced by \citet{shannon2001mathematical} also measures the level of uncertainty of samples obtained from a probability distribution, i.e. the randomness of the chance variable. Thus, the lower is the entropy, the more certain the model about its prediction. The Binary Entropy $H$ is used to evaluate the uncertainty of binary classification, i.e., the prediction can be only $0$ or $1$. Therefore, we applied the Binary Entropy to measure the uncertainty of each clinical note that it belongs to an ICD-9 code. The Binary Entropy $H(x_k)$ of an instance $k$ is defined in equation~\ref{eq:binEnt}, in which $p_{\theta}(y|x_k)$ corresponds to the posterior probability of instance $x_k$ to belongs to class $y$, that was produced by the model $\theta$~\cite{fu2013survey, shannon2001mathematical}. The closer the Binary Entropy measure is to $0$, the more confident the model is about that particular instance.

\begin{equation}\label{eq:binEnt}
	H(x_k) = - p_{\theta}(y|x_k) \log_2(p_{\theta}(y|x_k)) - (1-p_{\theta}(y|x_k)) \log_2((1-p_{\theta}(y|x_k)))
\end{equation}

As it is still a multi-label task, one challenge was how to combine the uncertainty measures obtained by each label. Then, we decided to apply the arithmetic mean and the mode operations to combine the uncertainty measure of an instance for all labels under analysis. In this scenario, we expected that, with the arithmetic mean operation, the instances selected are the ones that tend to be uncertain. While, with mode operation, the instances selected will be the ones that present uncertainty in the most labels since it returns the value that appears most often.

On the other hand, the correlation strategy selects the most informative instances based on semi-supervising algorithms. In this scenario, the clustering process is suitable for an acquisition function because it did not require the label information, grouping instances based on their similarity, which results in subsets of instances with related features. Thus, the selection of a few instances of each group can provide a good representation of the input space for the model to learn the probability distribution~\cite{demello2020shattering}. In this case, we applied the kmeans++ to cluster the clinical notes features, following by an instance selection inside each cluster. One challenge faced in this scenario is which distance metric should be used. We decided to start with Euclidean distance because it is the most typically applied. For instance selection, we evaluate three different methods: i) random selection, ii) select the ones closest to the centroid, and iii) select the ones closest to the border, i.e., most far from the centroid.

Finally, we attempt to combine both measures in two different ways, the Two-Stage (TS) metric and the Weighted Uncertainty Metric (WUM). In the TS method, the goal is to realize a previous selection on the overall instances in the set $D_u$ before applying the uncertainty metric. Thus, we applied the kmeans++ and randomly selected a defined number of instances of each cluster to explore different regions in the input space. Next, the uncertainty measure is employed to evaluate these instances selected. In the case of WUM, a similarity measure is applied as a weight for the relevance obtained with the uncertainty measure. Such a measure should be the average similarity over all instances in the set $D_u$.  However, due to the high computational cost, this distance is calculated between the instance and the centroid of the cluster~\cite{fu2013survey}. Thus, the instances are grouped using Kmeans++ as well, and the distance between the instance and the centroid is used as a weight for the uncertainty metric.

%--------------------------------------------------------------
\section{Experiments}
%--------------------------------------------------------------

The Medical Information Mart for Intensive Care (MIMIC-III) is a large medical database that contains more than a decade of information about admitted patients in critical care units at the Beth Israel Deaconess Medical Center in Boston, Massachusetts~\cite{johnson2016mimic}. MIMIC-III comprises different types of information, such as medications, observations, notes charted by care providers, procedure codes, diagnostic codes, and imaging reports. Patients in this database are de-identified, allowing their free access to international researchers under a data use agreement~\cite{johnson2016mimic}.

The clinical notes of the MIMIC-III database are divided into categories, in which the "discharge summary" is employed in this experiment. The goal of this task is to identify the ICD-9 code of each clinical note automatically. To tackle this task, we selected similar methods in the literature to apply active learning to, which include Random Forest, Logistic Regression, FNN, and CNN~\cite{huang2019empirical, nigam2016applying}.

Initially, the clinical notes are cleaned, in which stopwords, punctuation, and de-identifiers are removed, and all text is transformed to lower case. Next, the vectorization method is applied to convert the text into numeric vectors, in which TFIDF and Word2Vec are used as employed in the \citet{huang2019empirical}'s framework. As the vectorization did not require the label information, this process is conducted before the selection of instances.

In Active Learning, random selection is used as the baseline, so we used random selection and \cite{huang2019empirical}'s framework results as baselines, which will be called as the benchmark. The acquisition function used was Binary Entropy with mean and with mode, Least Confidence with mean and with mode, kmeans++ with the three variations (random, center, border), average linkage with three variations (random, center, border), and a two-stage strategy, including entropy and Least Confidence with mean combined with kmeans++. Such functions are applied to the feature space produced by the vectorization method.

As the MIMIC-III clinical notes are an imbalanced dataset, we are using the division provided by \citet{mullenbach2018explainable} to have the training, validation, and test sets, allowing the experiments to be compared. Therefore, the training set is used as the instances pool for the active learning selection, while the test set is used to measure the model performance after training in the selected samples. The number of clinical notes for each set according to division from \citet{mullenbach2018explainable} for each top ICD-9 code under analysis is presented in table~\ref{tb:mimicSummary}. It is worth mentioning that we do not consider clinical notes that do not have labels in the top list under analysis.

\begin{table}[ht]
	\centering
	\caption{Number of clinical notes per set division according to division from \citet{mullenbach2018explainable} for each top ICD-9 codes under analysis. The clinical notes that do not has these codes assignment are removed.}
	\begin{tabular}{c|c|c|c}
		Top k ICD-9 codes & Training set & Validation set & Test set \\
		\hline
		$10$ & $36,375$ & $1,364$ & $2,823$\\
		$50$ & $44,350$ & $1,544$ & $3,173$
		%$500$ & $47,242$ & $1,622$ & $3,354$
	\end{tabular}
	\label{tb:mimicSummary}
\end{table}

%--------------------------------------------------------------
\subsection{Setting up}
%--------------------------------------------------------------

The most parameters used in this experiments was the same that provide the better results reported in benchmark~\cite{huang2019empirical}. However, we modified the models to make them less complex, because AL has a small amount of data for the training process, and complex models may overfit the data. In the case of Random Forest, we used $30$ trees as the benchmark settings. However, we also set up the maximum depth as $500$. For Logistic Regression, we used the Limited-memory  Broyden–Fletcher–Goldfarb–Shanno algorithm (LBFGS) to estimate the parameters because it is adequate for large-scale problems, i.e., that contains many variables~\cite{liu1989limited}. Next, the architecture used for the FNN method is composed of $2$ hidden layers, having $500$ and $100$ neurons, respectively, with ReLU activation function. A sigmoid function was used in the output layer, in which the number of neurons is the same as the number of labels under analysis.

In each iteration of the experiments, $10$ samples were selected using a particular acquisition function, excepted for the first iteration in which random samples were selected. We set the maximum number of iterations to $300$, leading to a training set of $3000$ clinical notes in the last iteration. The evaluation for the training set used in the experiments was conducted using just the samples selected to train the model, while the evaluation for the test set was performed on the entire test set.

%--------------------------------------------------------------
\subsection{Top 10 ICD-9 codes}
%--------------------------------------------------------------

\begin{figure*}[ht]
	\centering
	\begin{subfigure}[h]{.49\textwidth}
		\includegraphics[width=\textwidth]{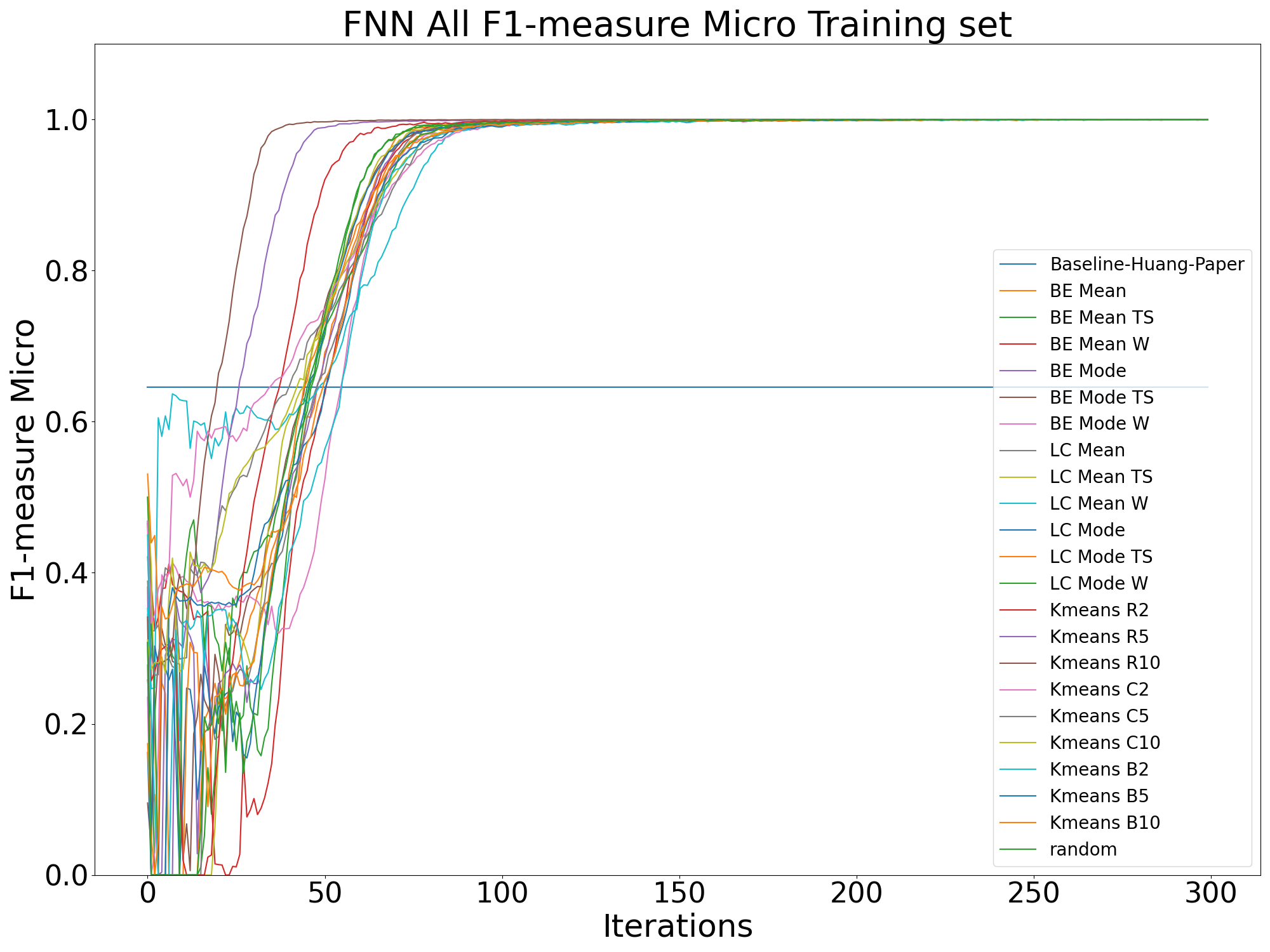}
		%\caption{Training set.}
	\end{subfigure}
	\begin{subfigure}[h]{.49\textwidth}
		\includegraphics[width=\textwidth]{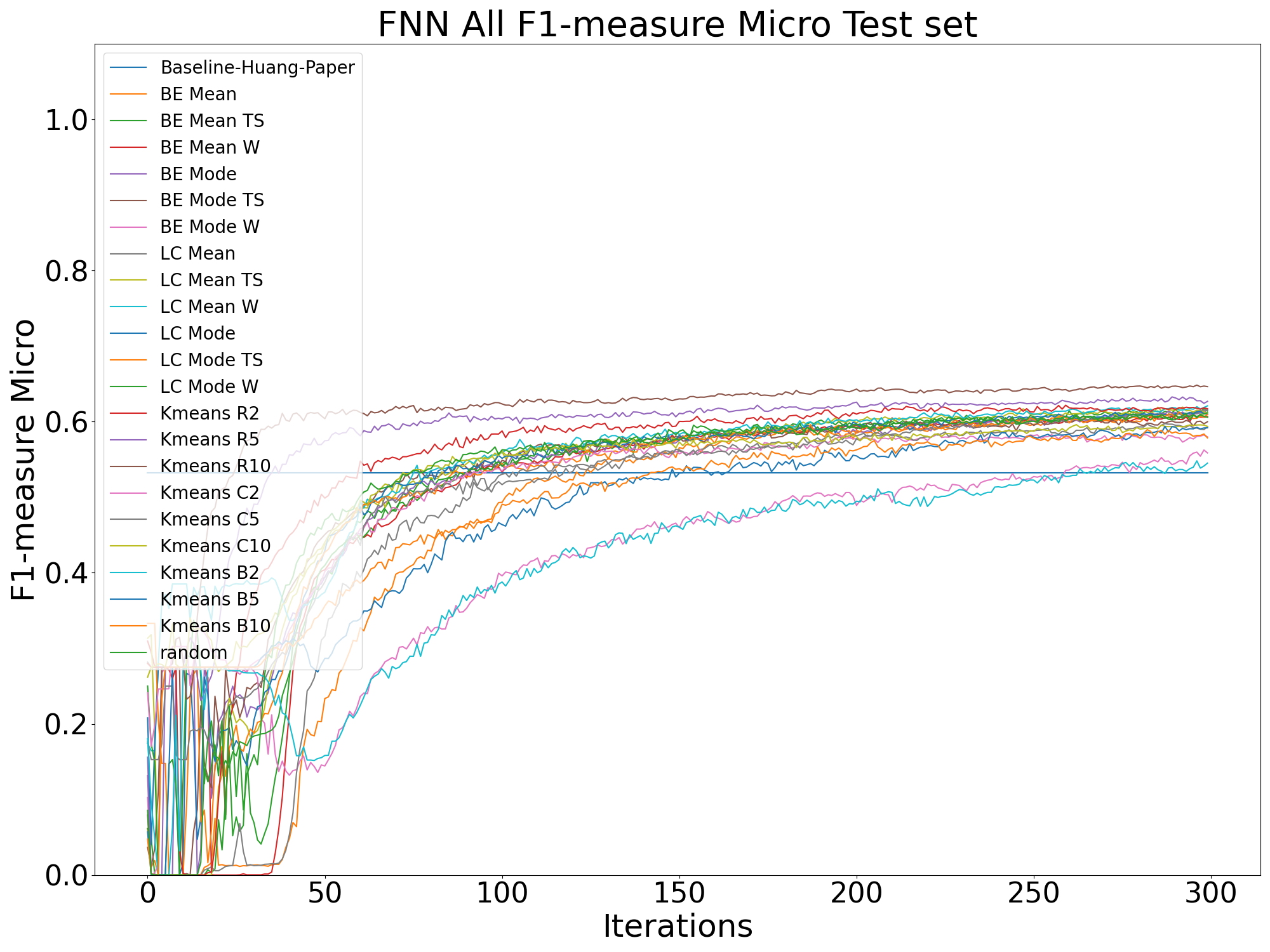}
		%\caption{Test set.}
	\end{subfigure}
	\caption{F1-measure (micro) results provided by FNN method combined with different acquisition functions for the top $10$ ICD-9 codes}
	\label{fig:dnnf1_micro}
\end{figure*}

Considering the top $10$ ICD-9 codes, we evaluated different acquisition functions using the Feed-forward Neural Network (FNN) to understand the behavior of each acquisition function in the feature space. The F1 measure (micro) for each acquisition function is presented in fig.~\ref{fig:dnnf1_micro}. All measures obtained is shown in table~\ref{tab:fnn10all} for $3000$ instances selected. All acquisition functions outperformed the results reported in the benchmark~\cite{huang2019empirical}. In this scenario, the uncertainty metrics provided lower or similar results than the random selection. Although the mean operation provided better results, the mode and mean operation do not result in a significant difference. We believe that this behavior is because the uncertainty points are too closed to the hyperplanes margin, representing the overlap and outliers points of each class, making the region definition for the class representation difficult.

On the other hand, some configurations using the clustering method outperform the random selection, improving the recall and precision measures. The selection of the samples farthest from the centroid (border) performed worse than the random and center selection. We believed that this occurs because the border samples are positioned in the overlap of the probabilities distributions of each label, which may represent an outlier, jeopardizing the model learning. The selection of the samples closest to the centroid (center) performed worse than the random selection. We believed that the center samples are a good representation of the local region, allowing the model convergence. However, these samples are not enough to provide a good representation of the probability distribution. These behaviors explain why the random selection points inside each cluster performed better because, with random selection, samples in the middle of the region could be selected, providing a good representation of the class region.

In summary, some acquisition functions with FNN outperformed the results reported in the benchmark~\cite{huang2019empirical} and the random selection. On the benchmark, they reported that FNN achieved an F1-measure (micro) of $0.53$, while the random selection achieved $0.61$ with $3000$ instances. In this scenario, the best result obtained was $0.65$ with kmeans++ ($10$ groups) with a random selection and $3000$ instances. We believed that the results were better than the benchmark because a most simple architecture combined with a suitable selection of the most informative points could better represent the input space, leading to better learning. Moreover, these results indicate that instances selection based on the features correlations is more suitable for FNN classifier because the uncertainty metric selection provided similar results with the random selection.

\begin{table}[ht!]
    \small
	\centering
	\caption{Measures obtained using different acquisition functions with FNN for coding assignment of top $10$ ICD-9 codes with $3000$ instances selected gradually. In this scenario, Prec. is Precision, F1 is the F1-measure, BE represents the Binary Entropy, LC is the Least Confidence, W corresponds to the weighted uncertainty metric, TS is the Two-Stage method, R represents the random selection, C corresponds to the selection of points closest to the centroid, and B is the selection of points far from the centroid. All representation is followed by the number of clusters analyzed.}
	\begin{tabular}{c|c|c|c|c|c|c}
		\multirow{2}{*}{AL strategies} & 
		\multicolumn{3}{c|}{Training} & \multicolumn{3}{c}{Test}\\
		& Prec. & Recall & F1 & Prec. & Recall & F1 \\
		\hline
		Random & $1.0$ & $1.0$ & $1.0$ & $0.66$ & $0.56$ & $0.61$\\
		Benchmark & $0.79$ & $0.57$ & $0.65$ & $0.68$ & $0.46$ & $0.53$\\
		\hline
		BE Mean & $1.0$ & $1.0$ & $1.0$ & $0.68$ & $0.55$ & $0.61$\\
        BE Mean TS & $1.0$ & $1.0$ & $1.0$ & $0.68$ & $0.56$ & $0.61$\\
        BE Mean W & $1.0$ & $1.0$ & $1.0$ & $0.68$ & $0.55$ & $0.61$\\
        BE Mode & $1.0$ & $1.0$ & $1.0$ & $0.67$ & $0.57$ & $0.62$\\
        BE Mode TS & $1.0$ & $1.0$ & $1.0$ & $0.67$ & $0.54$ & $0.6$\\
        BE Mode W & $1.0$ & $1.0$ & $1.0$ & $0.59$ & $0.53$ & $0.56$\\
        LC Mean & $1.0$ & $1.0$ & $1.0$ & $0.68$ & $0.56$ & $0.62$\\
        LC Mean TS & $1.0$ & $1.0$ & $1.0$ & $0.68$ & $0.55$ & $0.61$\\
        LC Mean W & $1.0$ & $1.0$ & $1.0$ & $0.69$ & $0.57$ & $0.62$\\
        LC Mode & $1.0$ & $1.0$ & $1.0$ & $0.68$ & $0.55$ & $0.61$\\
        LC Mode TS & $1.0$ & $1.0$ & $1.0$ & $0.67$ & $0.56$ & $0.61$\\
        LC Mode W & $1.0$ & $1.0$ & $1.0$ & $0.67$ & $0.56$ & $0.61$\\
        \hline
		Kmeans R2 & $1.0$ & $1.0$ & $1.0$ & $0.67$ & $0.58$ & $0.62$\\
        Kmeans R5 & $1.0$ & $1.0$ & $1.0$ & $0.64$ & $0.61$ & $0.63$\\
        Kmeans R10 & $1.0$ & $1.0$ & $1.0$ & $0.67$ & $0.63$ & $0.65$\\
        Kmeans C2 & $1.0$ & $1.0$ & $1.0$ & $0.57$ & $0.59$ & $0.58$\\
        Kmeans C5 & $1.0$ & $1.0$ & $1.0$ & $0.61$ & $0.57$ & $0.59$\\
        Kmeans C10 & $1.0$ & $1.0$ & $1.0$ & $0.6$ & $0.59$ & $0.59$\\
        Kmeans B2 & $1.0$ & $1.0$ & $1.0$ & $0.62$ & $0.49$ & $0.55$\\
        Kmeans B5 & $1.0$ & $1.0$ & $1.0$ & $0.62$ & $0.57$ & $0.59$\\
        Kmeans B10 & $1.0$ & $1.0$ & $1.0$ & $0.63$ & $0.53$ & $0.58$\\
	\end{tabular}
	\label{tab:fnn10all}
\end{table}

\begin{figure*}[ht]
	\centering
	\begin{subfigure}[h]{.49\textwidth}
		\includegraphics[width=\textwidth]{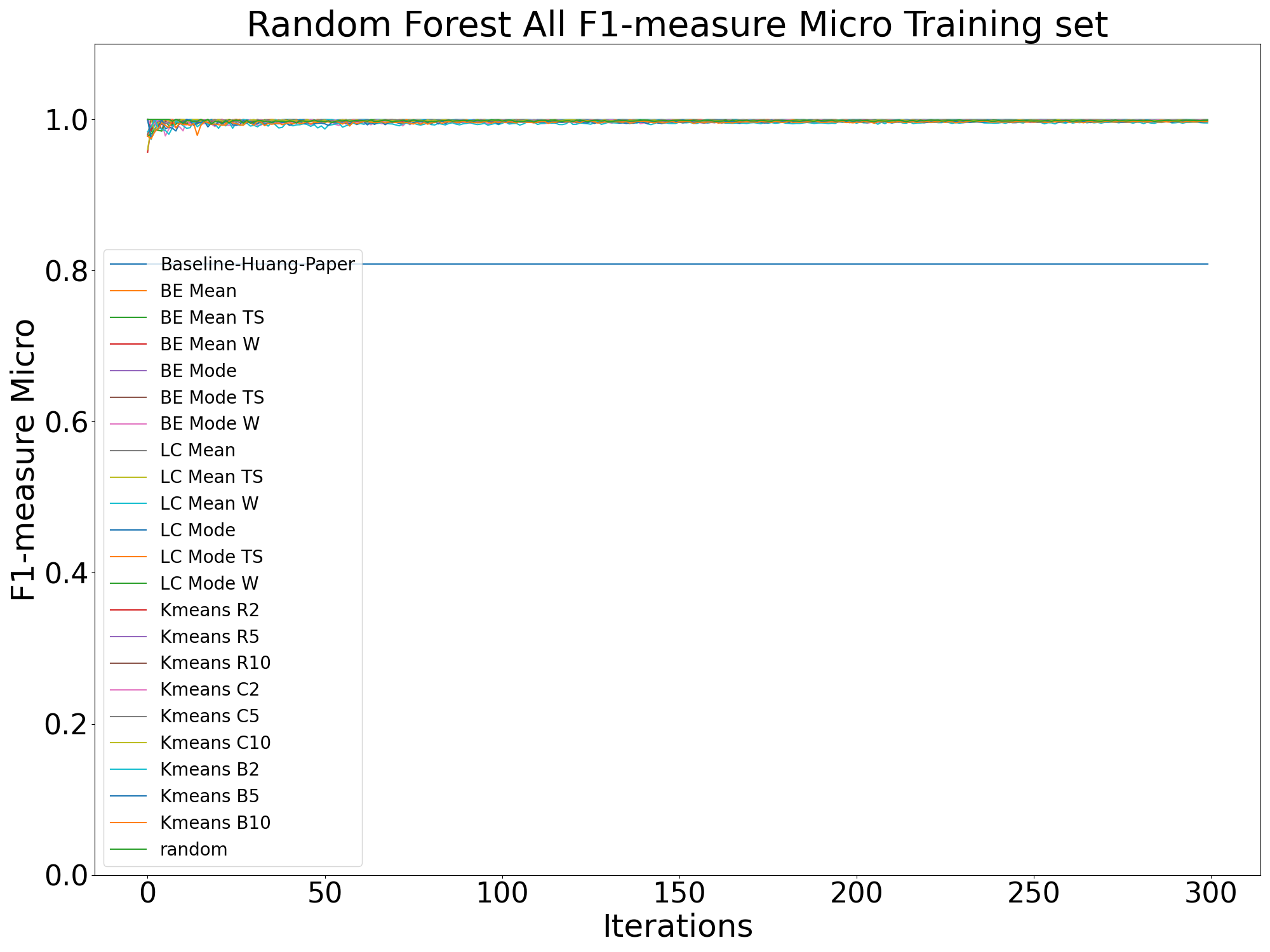}
		%\caption{Training set.}
	\end{subfigure}
	\begin{subfigure}[h]{.49\textwidth}
		\includegraphics[width=\textwidth]{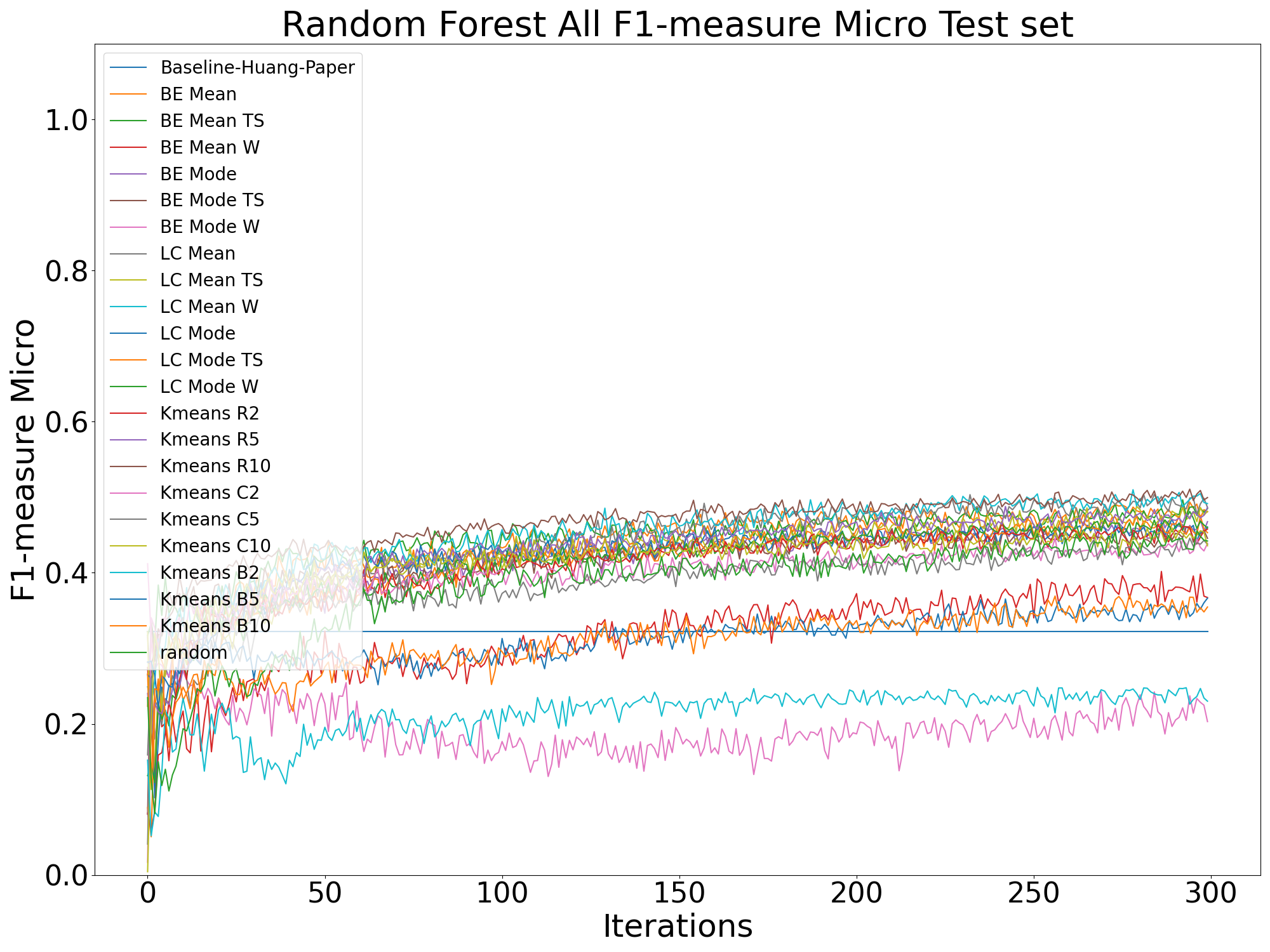}
		%\caption{Test set.}
	\end{subfigure}
	\caption{F1-measure (micro) results provided by Random Forest method combined with different acquisition functions for the top $10$ ICD-9 codes}
	\label{fig:rff1_micro}
\end{figure*}

Next, we did the same evaluation using Random Forest classification. The F1 measure (micro) for each acquisition function is presented in fig.~\ref{fig:rff1_micro}. All measures obtained is shown in table~\ref{tab:rf10all} for $3000$ instances selected. All acquisition functions outperformed the results reported in the benchmark~\cite{huang2019empirical}, in which mean operation provided a better recall measure, while mode operation provides a better precision measure. In this scenario, some acquisition functions with Random Forest outperformed the results reported in the benchmark~\cite{huang2019empirical} and the random selection. The benchmark reported that Random Forest achieved an F1-measure of $0.32$, while the random selection achieved $0.44$ with $3000$ instances. Thus, the best result obtained with AL strategies was $0.5$ using kmeans++ with random selection inside each cluster ($10$ groups) and $3000$ instances. We believed that the results reported were higher than the benchmark due to the definition of the depth parameter of the classifiers. Moreover, a suitable selection of the most informative instances for training set improved the representation of the probability distribution and improved the learning. In summary, the combination of both strategies is a good fit for the Random Forest classifier, although individual strategies also produce satisfactory results.

\begin{table}[ht!]
    \small
	\centering
	\caption{Measures obtained using different acquisition functions with Random Forest for coding assignment of top $10$ ICD-9 codes with $3000$ instances selected gradually. In this scenario, Prec. is Precision, F1 is the F1-measure, BE represents the Binary Entropy, LC is the Least Confidence, W corresponds to the weighted uncertainty metric, TS is the Two-Stage method, R represents the random selection, C corresponds to the selection of points closest to the centroid, and B is the selection of points far from the centroid. All representation is followed by the number of clusters analyzed.}
	\begin{tabular}{c|c|c|c|c|c|c}
		\multirow{2}{*}{AL strategies} & 
		\multicolumn{3}{c|}{Training} & \multicolumn{3}{c}{Test}\\
		& Prec. & Recall & F1 & Prec. & Recall & F1 \\
		\hline
		Random & $1.0$ & $1.0$ & $1.0$ & $0.66$ & $0.32$ & $0.44$\\
		Benchmark & $1.0$ & $0.7$ & $0.81$ & $0.76$ & $0.23$ & $0.32$\\
		\hline
		BE Mean & $1.0$ & $1.0$ & $1.0$ & $0.7$ & $0.37$ & $0.48$\\
        BE Mean TS & $1.0$ & $1.0$ & $1.0$ & $0.71$ & $0.37$ & $0.48$\\
        BE Mean W & $1.0$ & $1.0$ & $1.0$ & $0.7$ & $0.25$ & $0.37$\\
        BE Mode & $1.0$ & $1.0$ & $1.0$ & $0.76$ & $0.34$ & $0.47$\\
        BE Mode TS & $1.0$ & $1.0$ & $1.0$ & $0.74$ & $0.32$ & $0.45$\\
        BE Mode W & $1.0$ & $1.0$ & $1.0$ & $0.72$ & $0.12$ & $0.2$\\
        LC Mean & $1.0$ & $1.0$ & $1.0$ & $0.68$ & $0.38$ & $0.49$\\
        LC Mean TS & $1.0$ & $1.0$ & $1.0$ & $0.73$ & $0.36$ & $0.48$\\
        LC Mean W & $1.0$ & $1.0$ & $1.0$ & $0.7$ & $0.38$ & $0.49$\\
        LC Mode & $1.0$ & $1.0$ & $1.0$ & $0.74$ & $0.34$ & $0.46$\\
        LC Mode TS & $1.0$ & $1.0$ & $1.0$ & $0.73$ & $0.33$ & $0.45$\\
        LC Mode W & $1.0$ & $1.0$ & $1.0$ & $0.72$ & $0.34$ & $0.46$\\
        \hline
		Kmeans R2 & $1.0$ & $1.0$ & $1.0$ & $0.74$ & $0.33$ & $0.46$\\
        Kmeans R5 & $1.0$ & $1.0$ & $1.0$ & $0.78$ & $0.35$ & $0.49$\\
        Kmeans R10 & $1.0$ & $1.0$ & $1.0$ & $0.76$ & $0.37$ & $0.5$\\
        Kmeans C2 & $1.0$ & $1.0$ & $1.0$ & $0.66$ & $0.33$ & $0.44$\\
        Kmeans C5 & $1.0$ & $1.0$ & $1.0$ & $0.69$ & $0.32$ & $0.44$\\
        Kmeans C10 & $1.0$ & $1.0$ & $1.0$ & $0.67$ & $0.32$ & $0.44$\\
        Kmeans B2 & $1.0$ & $1.0$ & $1.0$ & $0.68$ & $0.14$ & $0.37$\\
        Kmeans B5 & $1.0$ & $1.0$ & $1.0$ & $0.69$ & $0.25$ & $0.35$\\
        Kmeans B10 & $1.0$ & $1.0$ & $1.0$ & $0.73$ & $0.24$ & $0.44$\\
	\end{tabular}
	\label{tab:rf10all}
\end{table}

\begin{figure*}[ht]
	\centering
	\begin{subfigure}[h]{.49\textwidth}
		\includegraphics[width=\textwidth]{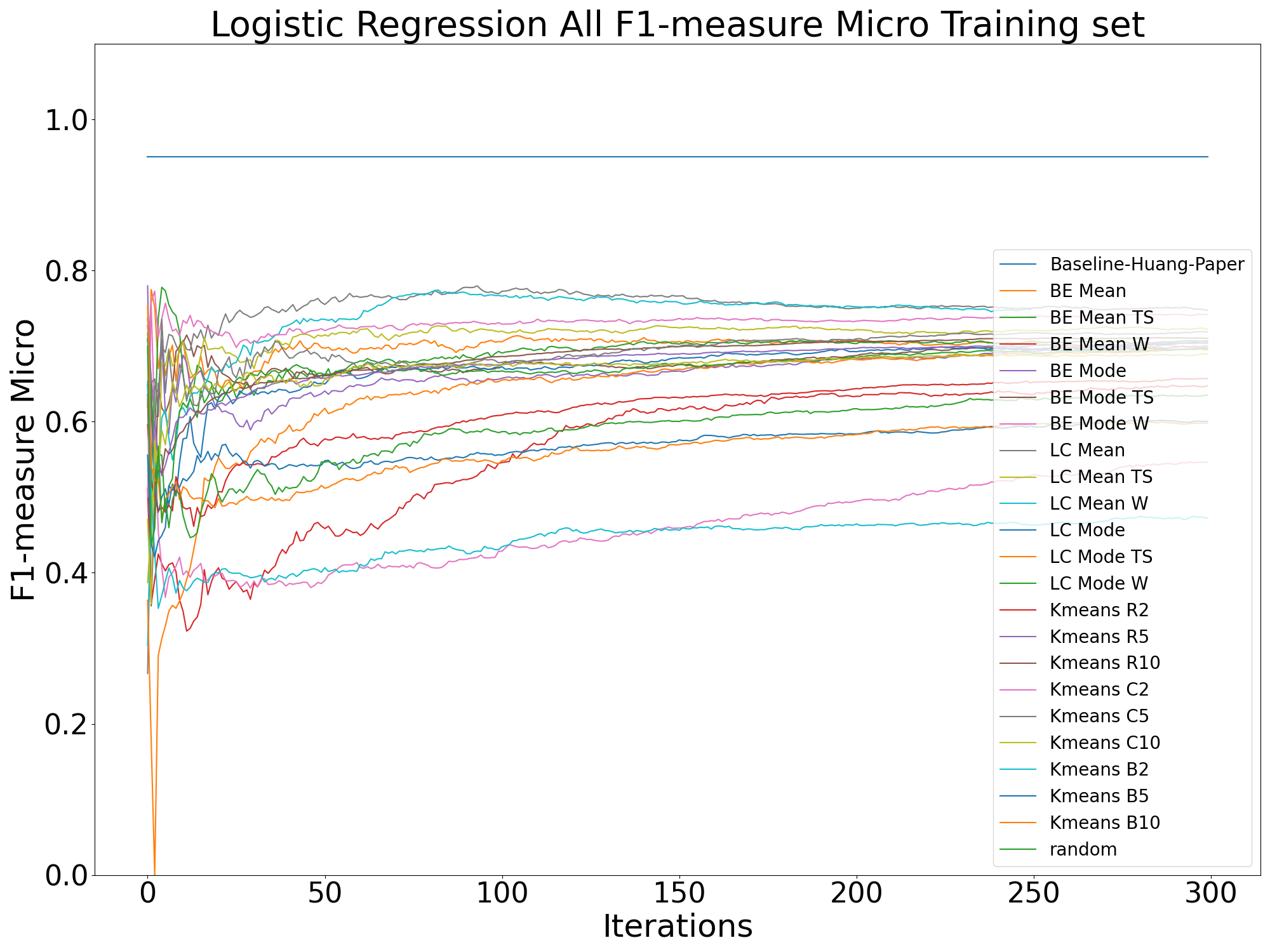}
		%\caption{Training set.}
	\end{subfigure}
	\begin{subfigure}[h]{.49\textwidth}
		\includegraphics[width=\textwidth]{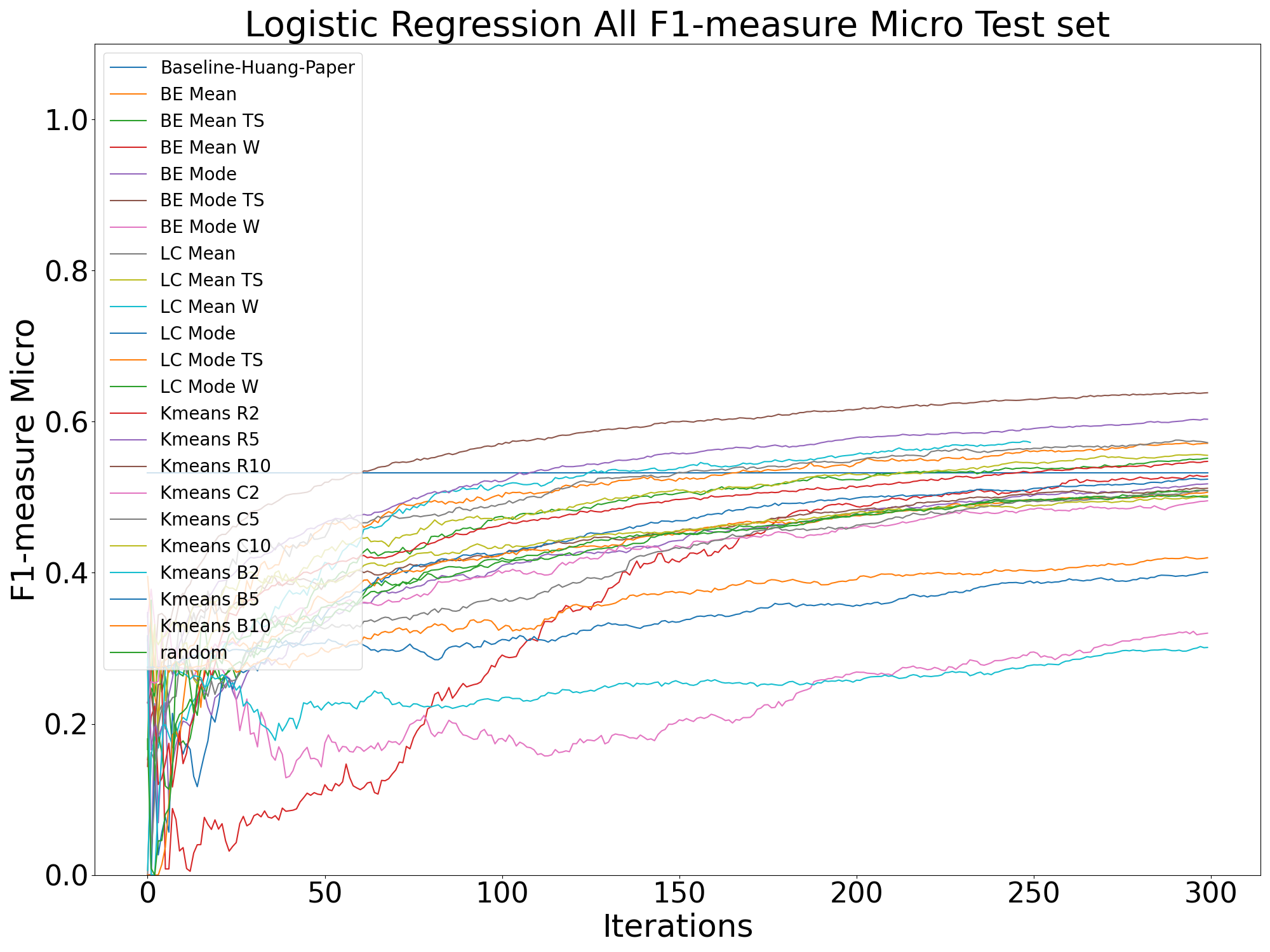}
		%\caption{Test set.}
	\end{subfigure}
	\caption{F1-measure (micro) results provided by Logistic Regression method combined with different acquisition functions for the top $10$ ICD-9 codes}
	\label{fig:lrf1_micro}
\end{figure*}

We also executed the same analysis using Logistic Regression, in which the F1 measure (micro) for each acquisition function is presented in fig.~\ref{fig:lrf1_micro}. All measures obtained is shown in table~\ref{tab:lr10all} for $3000$ instances selected. All acquisition functions outperformed the results reported in the benchmark~\cite{huang2019empirical}. In this scenario, some acquisition functions with Logistic regression outperformed the results reported in the benchmark~\cite{huang2019empirical} for the test set but provided lower measures on the training set, except in precision measurement. However, this behavior on the training set indicates that the Logistic Regression can provide a better generalization because the measure obtained on the training set is similar to the one obtained with the test set. In the benchmark, they reported that Logistic Regression achieved an F1-measure of $0.53$, while the random selection achieved $0.5$ with $3000$ instances. Thus, the best result obtained with AL strategies was $0.64$ using kmeans++ with random selection inside each cluster ($10$ groups) and $3000$ instances. Those results indicate that uncertainty metrics and features correlation can be a suitable AL strategy for this task. However, the combination of both did not improve the results.

\begin{table}[ht!]
    \small
	\centering
	\caption{Measures obtained using different acquisition functions with Logistic Regression for coding assignment of top $10$ ICD-9 codes with $3000$ instances selected gradually. In this scenario, Prec. is Precision, F1 is the F1-measure, BE represents the Binary Entropy, LC is the Least Confidence, W corresponds to the weighted uncertainty metric, TS is the Two-Stage method, R represents the random selection, C corresponds to the selection of points closest to the centroid, and B is the selection of points far from the centroid. All representation is followed by the number of clusters analyzed.}
	\begin{tabular}{c|c|c|c|c|c|c}
		\multirow{2}{*}{AL strategies} & 
		\multicolumn{3}{c|}{Training} & \multicolumn{3}{c}{Test}\\
		& Prec. & Recall & F1 & Prec. & Recall & F1 \\
		\hline
		Random & $0.9$ & $0.49$ & $0.64$ & $0.76$ & $0.37$ & $0.5$\\
		Benchmark & $0.96$ & $0.94$ & $0.95$ & $0.58$ & $0.49$ & $0.53$\\
		\hline
		BE Mean & $0.92$ & $0.56$ & $0.7$ & $0.73$ & $0.47$ & $0.57$\\
        BE Mean TS & $0.9$ & $0.58$ & $0.71$ & $0.74$ & $0.44$ & $0.55$\\
        BE Mean W & $0.93$ & $0.5$ & $0.65$ & $0.74$ & $0.41$ & $0.53$\\
        BE Mode & $0.92$ & $0.57$ & $0.7$ & $0.76$ & $0.39$ & $0.52$\\
        BE Mode TS & $0.91$ & $0.57$ & $0.7$ & $0.76$ & $0.39$ & $0.51$\\
        BE Mode W & $0.94$ & $0.38$ & $0.55$ & $0.74$ & $0.2$ & $0.32$\\
        LC Mean & $0.92$ & $0.63$ & $0.75$ & $0.73$ & $0.47$ & $0.57$\\
        LC Mean TS & $0.9$ & $0.6$ & $0.72$ & $0.74$ & $0.44$ & $0.56$\\
        LC Mean W & $0.92$ & $0.63$ & $0.69$ & $0.77$ & $0.47$ & $0.49$\\
        LC Mode & $0.91$ & $0.57$ & $0.7$ & $0.76$ & $0.4$ & $0.52$\\
        LC Mode TS & $0.91$ & $0.56$ & $0.7$ & $0.76$ & $0.38$ & $0.51$\\
        LC Mode W & $0.91$ & $0.56$ & $0.7$ & $0.76$ & $0.38$ & $0.51$\\
        \hline
		Kmeans R2 & $0.87$ & $0.53$ & $0.66$ & $0.77$ & $0.33$ & $0.55$\\
        Kmeans R5 & $0.86$ & $0.6$ & $0.7$ & $0.78$ & $0.35$ & $0.6$\\
        Kmeans R10 & $0.85$ & $0.61$ & $0.71$ & $0.79$ & $0.37$ & $0.64$\\
        Kmeans C2 & $0.87$ & $0.64$ & $0.74$ & $0.74$ & $0.33$ & $0.5$\\
        Kmeans C5 & $0.88$ & $0.61$ & $0.72$ & $0.74$ & $0.32$ & $0.51$\\
        Kmeans C10 & $0.86$ & $0.57$ & $0.69$ & $0.73$ & $0.32$ & $0.5$\\
        Kmeans B2 & $0.86$ & $0.33$ & $0.47$ & $0.67$ & $0.0$ & $0.3$\\
        Kmeans B5 & $0.87$ & $0.46$ & $0.6$ & $0.73$ & $0.0$ & $0.4$\\
        Kmeans B10 & $0.88$ & $0.45$ & $0.6$ & $0.75$ & $0.0$ & $0.42$\\
	\end{tabular}
	\label{tab:lr10all}
\end{table}

Finally, we evaluate the AL strategies combined with CNN, composed of one embedding layer, two convolution layers interposed by a max-pooling layer, and an output layer that applies sigmoid as the activation function. The convolution layers have $5 \times 5$ filter size, $128$ neurons, and ReLU as activation function, while the max-pooling layers built with a $5 \times 5$ filter size without a stride. The loss function used was the binary cross-entropy, while the optimizer was Adam. The word2vec matrix obtained with the entire training set of the MIMIC-III dataset was employed as the values for the embedding layer, which was not trained with the CNN parameters~\cite{huang2019empirical}.

According to the results, the AL strategies could not provide similar results as the benchmark and could not outperform the random selection, indicating that these methods are not appropriate to select representative instances for the CNN classifier. However, the Binary Entropy with the mean operation improved the precision when compared to random selection. In the benchmark, they reported that CNN achieved an F1-measure of $0.64$, while the random selection achieved $0.38$ with $3000$ samples. Thus, the best result obtained with AL strategies was $0.4$ using Binary Entropy with mean operation and $3000$ instances. However, these results was worse than the benchmark, and similar with random selection. Such behavior indicates that more studies are require to apply Al strategies with the CNN classifier, which produce an overfit for all configurations executed. In fact, this task requires a high complexity CNN model and, consequently, a large amount of instances.

\begin{figure*}[ht]
	\centering
	\begin{subfigure}[h]{.49\textwidth}
		\includegraphics[width=\textwidth]{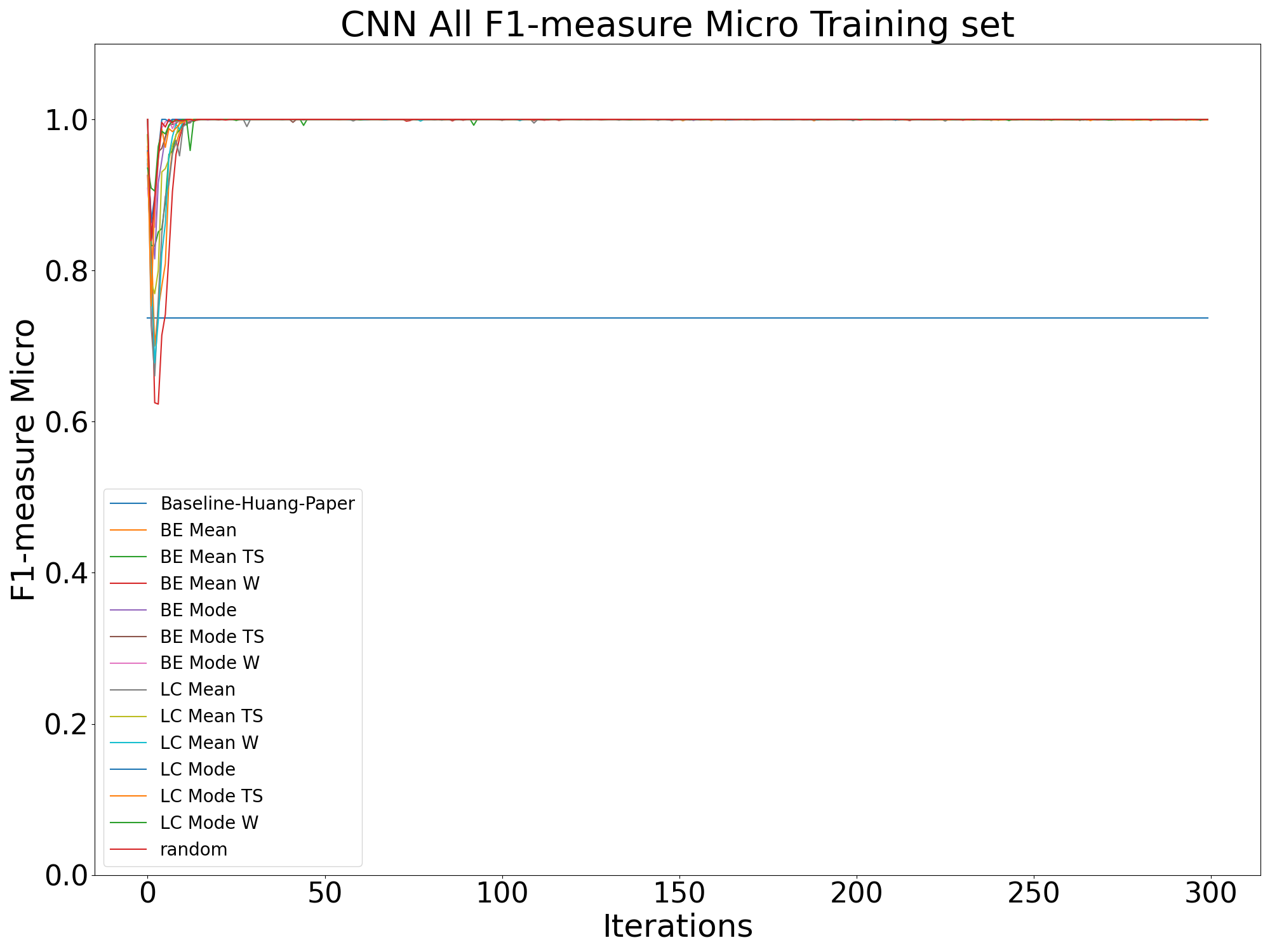}
		%\caption{Training set.}
	\end{subfigure}
	\begin{subfigure}[h]{.49\textwidth}
		\includegraphics[width=\textwidth]{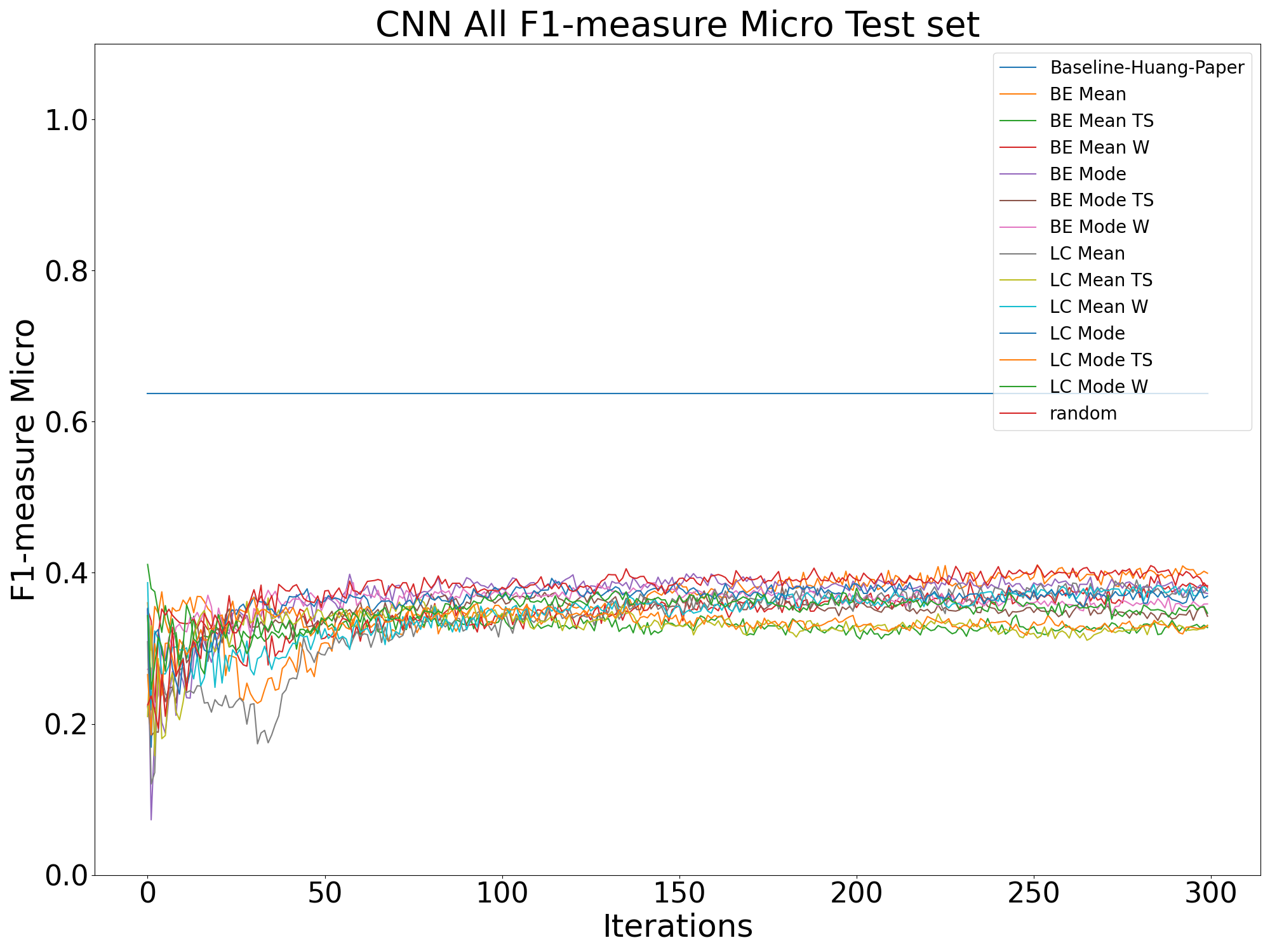}
		%\caption{Test set.}
	\end{subfigure}
	\caption{F1-measure (micro) results provided by CNN classifier combined with different acquisition functions for the top $10$ ICD-9 codes}
	\label{fig:cnnf1_micro}
\end{figure*}

\begin{table}[ht!]
    \small
	\centering
	\caption{Measures obtained using different acquisition functions with CNN using Word2Vec matrix in the embedding layer for coding assignment of top $10$ ICD-9 codes with $3000$ instances selected gradually. In this scenario, Prec. is Precision, F1 is the F1-measure, BE represents the Binary Entropy, LC is the Least Confidence, W corresponds to the weighted uncertainty metric, TS is the Two-Stage method, R represents the random selection, C corresponds to the selection of points closest to the centroid, and B is the selection of points far from the centroid. All representation is followed by the number of clusters analyzed.}
	\begin{tabular}{c|c|c|c|c|c|c}
		\multirow{2}{*}{AL strategies} & 
		\multicolumn{3}{c|}{Training} & \multicolumn{3}{c}{Test}\\
		& Prec. & Recall & F1 & Prec. & Recall & F1 \\
		\hline
		Random & $1.0$ & $1.0$ & $1.0$ & $0.41$ & $0.36$ & $0.38$\\
		Benchmark & $0.83$ & $0.67$ & $0.74$ & $0.74$ & $0.57$ & $0.64$\\
		\hline
		BE Mean & $1.0$ & $1.0$ & $1.0$ & $0.48$ & $0.34$ & $0.4$\\
        BE Mean TS & $1.0$ & $1.0$ & $1.0$ & $0.37$ & $0.29$ & $0.33$\\
        BE Mean W & $1.0$ & $1.0$ & $1.0$ & $0.45$ & $0.33$ & $0.38$\\
        BE Mode & $1.0$ & $1.0$ & $1.0$ & $0.41$ & $0.34$ & $0.37$\\
        BE Mode TS & $1.0$ & $1.0$ & $1.0$ & $0.38$ & $0.31$ & $0.34$\\
        BE Mode W & $1.0$ & $1.0$ & $1.0$ & $0.39$ & $0.33$ & $0.36$\\
        LC Mean & $1.0$ & $1.0$ & $1.0$ & $0.45$ & $0.33$ & $0.38$\\
        LC Mean TS & $1.0$ & $1.0$ & $1.0$ & $0.37$ & $0.3$ & $0.33$\\
        LC Mean W & $1.0$ & $1.0$ & $1.0$ & $0.43$ & $0.34$ & $0.38$\\
        LC Mode & $1.0$ & $1.0$ & $1.0$ & $0.4$ & $0.34$ & $0.37$\\
        LC Mode TS & $1.0$ & $1.0$ & $1.0$ & $0.37$ & $0.3$ & $0.33$\\
        LC Mode W & $1.0$ & $1.0$ & $1.0$ & $0.38$ & $0.32$ & $0.35$\\
	\end{tabular}
	\label{tab:cnn10all}
\end{table}

In summary, AL strategies combined with different classifiers provide good results for coding assignments with $10$ labels on the MIMIC-III database. This evaluation indicates that each classifier requires a different instance selection due to its particular preference bias to provide a good model of the data. The benchmark best F1-measure reported was $0.7$ using a Recurrence Neural Network with GRU units ($36,375$ instances). However, with the classifier employed in this evaluation, the best results reported were $0.64$ with CNN ($36,375$ instances), while the best results achieved with AL strategies were $0.64$ using Logistic Regression and kmeans++ with random selection inside each cluster ($3000$ instances).

%--------------------------------------------------------------
\subsection{Applying the best configurations}
%--------------------------------------------------------------

Considering the top $50$ ICD-9 codes, we executed just the random selection as a baseline and the respective configuration that provided the best results, which was kmeans++ with random selection inside each cluster ($10$ groups), outperforming the random selection. All measures obtained is shown in table~\ref{tab:all50} for $3000$ instances selected, except the CNN because it did not provide good results with the AL strategies employed in those experiments.

In the case of the FNN classifier, all configurations executed outperformed the results reported in the benchmark~\cite{huang2019empirical}, however only the clustering methods as acquisition function provided better results than the random selection. In the Random Forest classifier, all configurations executed outperformed or produced similar results like the ones reported in the benchmark~\cite{huang2019empirical} and the random selection. The Logistic Regression combined with AL strategy also produced better results compared with the random selection and the benchmark~\cite{huang2019empirical}. It is worth mentioning that Logistic Regression requires a set of instances representing all labels, so this algorithm starts with a random selection of $50$ samples that represents all labels.

\begin{table}[ht!]
    \small
	\centering
	\caption{F1 measure (micro) for coding assignment of top $50$ ICD-9 codes.}
	\begin{tabular}{c|c|c|c|c|c|c|c}
	    \multirow{2}{*}{Classifier} & \multirow{2}{*}{AL Strategies} & \multicolumn{6}{c}{\#Samples}\\
		&& $500$ & $1000$ & $1500$ & $2000$ & $2500$ & $3000$ \\
		\hline
		\multirow{2}{*}{FNN} &
		Random & $0.0$ & $0.17$ & $0.27$ & $0.34$ & $0.38$ & $0.41$\\
		& Kmeans R10 & $0.26$ & $0.43$ & $0.47$ & $0.48$ & $0.48$ & $0.49$\\
		\hline
		\multirow{2}{*}{RF} &
		Random & $0.13$ & $0.16$ & $0.16$ & $0.16$ & $0.19$ & $0.18$\\
		& Kmeans R10 & $0.19$ & $0.22$ & $0.24$ & $0.23$ & $0.24$ & $0.26$\\
		\hline
		\multirow{2}{*}{LR} &
		Random & $0.22$ & $0.29$ & $0.32$ & $0.35$ & $0.37$ & $0.38$\\
		& Kmeans R10 & $0.26$ & $0.33$ & $0.37$ & $0.4$ & $0.41$ & $0.43$\\
	\end{tabular}
	\label{tab:all50iter}
\end{table}

\begin{table}[ht!]
    \small
	\centering
	\caption{F1 measure (micro) for coding assignment of top $50$ ICD-9 codes.}
	\begin{tabular}{c|c|c|c|c|c|c|c}
		\multirow{2}{*}{} & \multirow{2}{*}{AL strategies} & 
		\multicolumn{3}{c|}{Training} & \multicolumn{3}{c}{Test}\\
		&& Prec. & Recall & F1 & Prec. & Recall & F1 \\
		\hline
		\multirow{3}{*}{FNN} &
		Benchmark & $0.25$ & $0.11$ & $0.13$ & $0.23$ & $0.11$ & $0.12$\\
		& Random & $0.95$ & $0.74$ & $0.84$ & $0.63$ & $0.27$ & $0.38$\\
		& Kmeans R10 & $0.99$ & $0.98$ & $0.99$ & $0.53$ & $0.45$ & $0.48$\\
		\hline
		\multirow{3}{*}{RF} &
		Benchmark & $1.0$ & $0.29$ & $0.39$ & $0.54$ & $0.1$ & $0.12$\\
		& Random & $1.0$ & $0.98$ & $0.99$ & $0.77$ & $0.1$ & $0.18$\\
		& Kmeans R10 & $1.0$ & $0.99$ & $0.99$ & $0.8$ & $0.15$ & $0.26$\\
		\hline
		\multirow{3}{*}{LR} &
		Benchmark & $0.99$ & $0.98$ & $0.98$ & $0.44$ & $0.32$ & $0.37$\\
		& Random & $0.86$ & $0.35$ & $0.5$ & $0.78$ & $0.25$ & $0.38$\\
		& Kmeans R10 & $0.84$ & $0.41$ & $0.55$ & $0.78$ & $0.3$ & $0.43$\\
	\end{tabular}
	\label{tab:all50}
\end{table}

% Considering the top $500$ ICD-9 codes, there are no results reported in the benchmark~\cite{huang2019empirical}, however we decided to analyze when the label space increases more. Then, we executed...

%--------------------------------------------------------------
\section{Conclusions}
%--------------------------------------------------------------

In this research, we evaluate the possibility of employing the most common active learning methods in the medical domain, focusing on coding assignment the clinical notes under the discharge summaries category. Typically, the data in this scenario is imbalanced due to the domain aspects, and, as the notes are manually annotated, the cost of coding the data is too high~\cite{mullenbach2018explainable, huang2019empirical}. Such aspects motivate the use of Active Learning methods because it can select the most informative samples to be labeled, reducing the annotation cost, and provide a better data balance for the training set.

According to the results, AL methods combined with the classifier models produced similar results to those reported in the literature, indicating that the selection of the most informative samples provides a good representation of the overall training set. The baseline for AL methods is random selection, which was outperformed by most of the techniques employed for different classifiers. Such results confirmed that AL methods could be used in the medical domain for coding assignment, keeping the results already achieved, and reducing the annotation cost because just a few documents have to be manually labeled. Besides, we noticed that using mode operation to combine the label information can improve the precision measure, while the mean operation enhances the recall. Thus, according to the scenario, this can be modified to satisfy the goals of the task under analysis.

One limitation of the application of AL is that the analysis of every unlabeled instance can be time-consuming. Besides, each model has a different search bias, indicating that other DL models may provide better results with a different approach. As future work, studies on AL with Bert models for coding assignment are required, once these models present the best results when dealing with the MIMIC-III database in a high dimensional labels space~\cite{mullenbach2018explainable, dor2020active}. However, as more complex is the model, more labeled instances are required to ensure the learning. Moreover, a better analysis of the clustering algorithms should be conducted, once the kmeans++ provided better results for the most scenarios analyzed. In this context, different clustering algorithms and distance metrics should be evaluated to improve the results produced by AL methods. Finally, visual analysis to interpret the model behavior or continuous learning to deal with situations that do not have all the label information is a possible mechanism to improve the medical domain tasks.

\section{Acknowledgments}

This work was supported by Mitacs Accelerate Program and the Semantic Health Company as the partner organization.

\bibliographystyle{ACM-Reference-Format}
\bibliography{references}

\end{document}